\documentclass[10pt, a4paper]{article}
\usepackage{lrec2022} 
\usepackage{multibib}
\newcites{languageresource}{Language Resources}
\usepackage{graphicx}
\usepackage{tabularx}
\usepackage{soul}
\usepackage{titlesec}
\titleformat{\section}{\normalfont\large\bfseries\center}{\thesection.}{1em}{}
\titleformat{\subsection}{\normalfont\SmallTitleFont\bfseries\raggedright}{\thesubsection.}{1em}{}
\titleformat{\subsubsection}{\normalfont\normalsize\bfseries\raggedright}{\thesubsubsection.}{1em}{}
\renewcommand\thesection{\arabic{section}}
\renewcommand\thesubsection{\thesection.\arabic{subsection}}
\renewcommand\thesubsubsection{\thesubsection.\arabic{subsubsection}}

\usepackage{epstopdf}
\usepackage[utf8]{inputenc}

\usepackage{hyperref}
\usepackage{xstring}

\usepackage{url}
\usepackage{breakurl}

\usepackage{color}

\title{Speech Resources in the Tamasheq Language}

\name{Marcely Zanon Boito¹, Fethi Bougares², Florentin Barbier³, Souhir Gahbiche³,\\ \large\textbf{Lo{\"i}c Barrault², Mickael Rouvier¹, Yannick Estève¹}} 

\address{¹LIA - Avignon University, France\\ 
        ²LIUM - Le Mans University, France\\
        ³Airbus - France\\
         \textbf{contact:} \{marcely.zanon-boito, yannick.esteve\}@univ-avignon.fr\\}

\abstract{
In this paper we present two datasets for Tamasheq, a developing language mainly spoken in Mali and Niger. These two datasets were made available for the IWSLT 2022 low-resource speech translation track, and they consist of collections of radio recordings from daily broadcast news in Niger~(Studio Kalangou) and Mali~(Studio Tamani).
We share (i) a massive amount of unlabeled audio data~(671 hours) in five languages: French from Niger, Fulfulde, Hausa, Tamasheq and Zarma, and (ii) a smaller 17 hours parallel corpus of audio recordings in Tamasheq, 
with utterance-level translations in the French language. All this data is shared under the Creative Commons BY-NC-ND 3.0 license. We hope these resources will inspire the speech community to develop and benchmark models using the Tamasheq language.
 \\\newline \Keywords{speech corpus, speech translation, tamasheq, zarma, hausa, fulfulde, french}
}

\begin{document}

\maketitleabstract

\section{Introduction}

The vast majority of speech pipelines are developed for and in \textit{high-resource} languages, a small percentage of languages for which there is 
a large amount of annotated data freely available~\cite{joshi2020state}. This 
not only limits the investigation of language impact in current pipelines, as the applied languages are usually from the same subset, but it also fails to reflect the real-world performance these approaches will have in diverse and smaller datasets.

In recent years, the IWSLT evaluation campaign\footnote{\url{https://iwslt.org/2022/low-resource}} introduced a low-resource speech translation track focused on developing and benchmarking translation tools for under-resourced languages. While for a vast majority of these languages, 
there is not enough parallel data for training large translation models,
in these cases we might still have access to limited disparate resources, such as word-level translations, small parallel textual data, monolingual texts and recordings. This track of the 
IWSLT campaign thus focuses on 
leveraging these different kinds of data for building 
effective translation systems under 
realistic settings.

In this paper we present the resources in the Tamasheq language we share in the context of the IWSLT 2022: low-resource speech translation track. 
Tamasheq is a variety of Tuareg, a Berber macro-language spoken by nomadic tribes across North Africa in Algeria, Mali, Niger and Burkina Faso~\citelanguageresource{heath2006dictionnaire}. 
It accounts for approximately 500,000 native speakers, being mostly spoken in 
Mali and Niger~\cite{ethnologue:tamasheq07122021}.

We share a large audio corpus, made of 224 hours of Tamasheq, together with 417 hours in other four languages of Niger~(French, Fulfulde, Hausa and Zarma). We also share a smaller corpus of 17 hours of Tamasheq utterances aligned with French translations. We hope that these resources will represent an interesting use-case for the speech community, allowing them to not only develop low-resource speech systems in Tamasheq, but also to investigate the leveraging of unannotated audio data in diverse languages that co-exist in the same geographic region. 

This paper is organized as follows. Section~\ref{sec:sourcecontent} presents the source content of the data shared: thanks to the \textit{Fondation Hirondelle Initiative} and local partners, we are able to collect broadcast news in diverse African languages. 
Section~\ref{tamasheqtrad} presents the small Tamasheq-French parallel corpus, and Section~\ref{rawaudiocollection} presents the collection of unannotated audio data in French from Niger, Fulfulde, Hausa, Tamasheq and Zarma. Finally, Section~\ref{baseline} presents a speech translation baseline model for the IWSLT 2022 campaign, and Section~\ref{conclusion} concludes this work.

\section{The source content: The Fondation Hirondelle Initiative}\label{sec:sourcecontent}

The Fondation Hirondelle\footnote{\url{https://www.hirondelle.org/en/}} is a Swiss non-profit organization founded in 1995 by journalists, with the goal of supporting local independent media in areas of social unrest. They produce and broadcast information and talk shows 
in different countries, providing local partners with editorial, managerial and structural support and training to function in a sustainable manner. 

In this work we focus on their daily radio broadcasts episodes, 
produced and broadcast by local partners 
in different languages. These allow the local communities to get informed in their own dialects, in contrast to mainstream media that tends to cover only the countries' official languages. For the Tamasheq language, we find these episodes being produced daily in Mali~(Studio Tamani\footnote{\url{https://www.studiotamani.org/}}) and Niger~(Studio Kalangou\footnote{\url{https://www.studiokalangou.org/}}).

\paragraph{Speech Style and Quality.} The radio episodes are recorded in local studios: for each episode, one or two hosts present the news, and often interviews and advertisements are included. Most of the speech is of good quality, 
with rare instances of background music during advertisements. For interviews, we notice some cases of overlapping speech, mainly when simultaneous translation is performed, 
and background noise such as outdoor sounds.

\paragraph{Audio Web-crawling.} 
With the authorization of the Fondation Hirondelle and partners, we downloaded the \texttt{.mp3} episodes by generating URLs from the local partners' 
broadcast webpages.\footnote{ \texttt{http://\textbf{<studio-name>}.org/journaux/}
} The corpora presented in Section~\ref{tamasheqtrad} and Section~\ref{rawaudiocollection} use these audio 
files as source content.

\section{The Tamasheq-French \\Parallel Corpus}\label{tamasheqtrad}

This corpus corresponds to 17 hours of \textit{controlled} speech utterances, with manual translations to the French language. We also share a 
longer version of this corpus, including 2 additional hours of potentially noisy speech segments. We detail below the steps for creating this corpus, and present 
general statistics.

\paragraph{1. Data Downloading.} 
100 
episodes were downloaded from the Studio Kalangou website in February 2019: 23 episodes from 2016, 36 episodes from 2017 and 2018, and 5 episodes from 2019. This results in 25 hours of raw speech, with an average episode duration of 15~minutes. 

\paragraph{2. Translation Process.} We commissioned ELDA~(Evaluations and Language resources Distribution Agency)\footnote{\url{http://www.elra.info/en/about/elda/}} for 
translating the 25 hours of Tamasheq into French text. No transcriptions were commissioned. 
The translations were produced by at least two native Tamasheq speakers,\footnote{The number and identity of the translators was not disclosed to us.} with posterior text correction by proficient French speakers. The translators had access to 5 pages of guidelines, 
including segmentation guidelines for slicing 
the episodes 
into utterances. Annotation used the \textit{Transcriber} open-source tool.\footnote{\url{http://trans.sourceforge.net/}} Lastly, some utterances contain gender annotation and speaker identification. Unfortunately, this annotation was not standardized across the different translators, and therefore some speakers are referred by different speaker ids in different files, with difficult disambiguation. We thus caution to the use of this information, as the current speaker ids might represent an upper-bound over the real number of speakers in this dataset.

\paragraph{3. Translation Post-processing.} From the original \textit{Transcriber} annotation files, we filtered out segments corresponding to pauses, noise and music, and removed segments flagged by the annotators as corresponding to foreign languages, such as Arabic and French.
We then applied \texttt{sacremoses}, the python port of the Moses toolkit~\cite{koehn2007moses}, for punctuation normalization and tokenization in French. During post-processing, we noticed that some segments~(roughly two hours) were flagged as being of poor source audio quality. For these, the translation was produced nevertheless, so we decided to include them 
in a larger \textit{less controlled} version of the shared corpus. 

\paragraph{4. Audio Post-processing.} We use the 
resulting collection of segments described in 3. to 
split the episodes into utterance-level audio files. For posterior use in standard speech processing libraries, we also convert the original \texttt{.mp3} files into \texttt{.wav}, 16bits, 16KHz, single channel. We then remove all utterances shorter than 1s and longer than 30s. This is the same audio pre-processing from~\newcite{baevski2020wav2vec}.

\paragraph{5. Statistics.} Table~\ref{tab:tmqfrastats} presents the statistics for the two versions of the Tamasheq-French parallel corpus we share with the community. The difference between \textit{clean}~(17\,h) and \textit{full}~(19\,h) 
is that the latter includes 
potentially noisy segments. 
Both 
are 
available through \texttt{GitHub}: \url{https://github.com/mzboito/IWSLT2022_Tamasheq_data}. 

Regarding the gender distribution, we notice that almost all labeled utterances correspond to male speech. We also observe that more than a half of the utterances are unlabeled~(\textit{unknown}). 
For having a better idea of the gender distribution for this dataset, we 
performed gender labeling 
using the \texttt{LIUM\_SpkDiarization} tool~\cite{meignier2010lium}. The results should be interpreted as an estimation, but we observed that all the unlabeled utterances seemed to belong to the male category. We thus believe that this dataset is unfortunately very gender unbalanced.

\begin{table*}
\centering
\resizebox{\textwidth}{!}{
\begin{tabular}{lcccc|c|cccc} \hline
                          & \multicolumn{4}{c|}{\textbf{clean (17\,h)}}                                 & \textbf{} & \multicolumn{4}{c}{\textbf{full (19\,h)}}                                 \\\
                  & \textbf{male} & \textbf{female} & \textbf{unknown} & \textbf{total} & \textbf{} & \textbf{male} & \textbf{female} & \textbf{unknown} & \textbf{total} \\\hline
\textbf{\# utterances}    & 2,313             & 10                & 3,506             & \textbf{5,829}        &           & 2,643             & 11                & 3,625             & \textbf{6,279}        \\
\textbf{duration}      & 07:37:54           & 0:00:48           & 10:04:49          & \textbf{17:43:33}     &           & 08:49:11           & 0:00:51           & 10:28:42          & \textbf{19:18:45}     \\
\hline
\end{tabular}}
\caption{Statistics for the clean~(left) and full~(right) Tamasheq-French parallel corpus.}
\label{tab:tmqfrastats}
\end{table*}

\section{The Niger-Mali Audio Collection}\label{rawaudiocollection}

This unannotated audio collection corresponds to 671~hours of episodes in five languages: French from Niger, Fulfulde, Hausa, Tamasheq and Zarma. We automatically segmented this audio data, 
generating 641~hours of content ready for deployment in speech processing pipelines. We detail below the creation of this audio collection, and present some general statistics. 

\begin{table*}
\centering
\resizebox{\textwidth}{!}{
\begin{tabular}{lcc|c|cccc}\hline
\textbf{}                          & \multicolumn{1}{l}{\textbf{\# episodes}} & \textbf{duration}   & \multicolumn{1}{l|}{} & \multicolumn{1}{l}{\textbf{\# utterances}} & \multicolumn{1}{l}{\textbf{duration (male)}} & \multicolumn{1}{l}{\textbf{duration (female)}} & \multicolumn{1}{l}{\textbf{duration (total)}} \\ \hline
\textbf{French}                    & 464                                      & 116:22:09           &                      & 38,332                                     & 52:15:07                                     & 58:46:00                                       & 111:01:07                                     \\
\textbf{Fulfulde}                   & 459                                      & 114:23:40           &                      & 39,255                                     & 73:31:36                                     & 35:54:47                                       & 109:26:23                                     \\
\textbf{Hausa}                     & 424                                      & 105:32:48           &                      & 35,684                                     & 75:05:12                                     & 25:39:40                                       & 100:44:52                                     \\
\textbf{Tamasheq}                  & 1,014                                    & 234:36:29           &                      & 75,995                                     & 134:11:32                                    & 90:33:44                                       & 224:45:16                                     \\
\textbf{Zarma}                     & 405                                      & 100:42:34           &                      & 34,198                                     & 57:03:37                                     & 38:55:33                                       & 95:59:10                                      \\ \hline
\multicolumn{1}{l}{\textbf{Total}} & \textbf{2,766}                           & \textbf{671:37:43} &                      & \textbf{223,464}                           & \textbf{392:07:04}                           & \textbf{249:49:44}                             & \textbf{641:56:50}     \\ \hline                      
\end{tabular}}
\caption{Statistics for the Niger-Mali audio collection raw content~(left) and automatically segmented version~(right), produced by the use of a speech segmentation system with gender labeling.}
\label{table:tamanikalangoustats}
\end{table*}

\paragraph{1. Data Downloading.} Similarly to Section~\ref{tamasheqtrad}, we downloaded 606 episodes in Tamasheq from Studio Tamani,\footnote{For Studio Tamani news are broadcast twice a day. These correspond to \textit{matin} and \textit{soir} segments in the source files, respectively morning and evening shows.} and 2,160 episodes in all the available languages from Studio Kalangou: French~(464), Fulfulde~(459), Hausa~(424), Tamasheq~(408) and Zarma~(405). These episodes correspond to the content we managed to retrieve with our web-crawler.\footnote{Accessing and downloading date: 07/10/2021.} It explored URLs ranging from November 2019 to September 2021 for Studio Kalangou, and from January 2020 to September 2021 for Studio Tamani.\footnote{Since the sites vary in their file indexing, not all episodes from the indicated periods are successfully retrieved.} The left portion of Table~\ref{table:tamanikalangoustats} presents the statistics for the downloaded \texttt{.mp3} files: 671\,h of audio, being 116\,h in French, 114\,h in Fulfulde, 105\,h in Hausa, 234\,h in Tamasheq, and 100\,h in Zarma. This corresponds to a total of 2,766 episodes, with an average episode duration of 15 minutes for files from Studio Kalangou, and 13 minutes for Studio Tamani. Finally, we highlight that the choice of having more audio data in Tamasheq was deliberated, since in this paper we focus on building resources for the Tamasheq language.


\paragraph{2. Segmenting Episodes into Breath Turns.}
The 
episodes downloaded from the websites are used as input for the \texttt{LIUM\_SpkDiarization} tool. 
The goal of this step is (i)~to produce a format 
compatible with current speech processing models, that cannot deal with very long speech turns, and (ii)~to remove silence, music and other non speech events. The \texttt{LIUM\_SpkDiarization} performs speech diarization, separating turns of speech. This allows us to slice the 
episodes into smaller audio chunks~(breath turns).\footnote{By default, the maximum turn length is set to 20 milliseconds.} It also has the advantage of producing gender annotation, which allow us to estimate the gender distribution for each language. After applying this diarization tool, we 
remove the first 12 seconds of each episode, as these often corresponded to intro jingles. The right portion of Table~\ref{table:tamanikalangoustats} presents the obtained result: 641\,h of audio, being 111\,h of French, 109\,h of Fulfulde, 100\,h of Hausa, 224\,h of Tamasheq, and 95\,h of Zarma. There are 392\,h estimated to be from male speakers, and 249\,h from female speakers.

\paragraph{3. Resulting Corpus.}
We make both versions of this corpus~(Table~\ref{table:tamanikalangoustats}) available to the community: the 671\,h corpus based on episodes, and the 641\,h version based on breath turns. This is because, even though we believe our segmentation process to be of good quality, it is still supported by an automatic diarization tool. By providing the source content, we allow the community to choose their own segmentation approach. The audio collection is made available through a dedicated website: \url{https://demo-lia.univ-avignon.fr/studios-tamani-kalangou/}. In the next section, we briefly elaborate on the five languages available. 

\subsection{The Languages}

The speech resources we collect and share in this paper correspond to five languages spoken in Niger: French, Fulfulde, Hausa, Tamasheq and Zarma. We now provide a brief description of these languages.

\begin{itemize}
    \item \textbf{French~(FRA):} French is the official language of the Niger, and a high-resource romance language from the indo-european family. At first, we intended to include only the other four languages listed in this section in the audio collection. However, we noticed some french segments in the Tamasheq annotation from Section~\ref{tamasheqtrad}, and hypothesized that some lexical borrowing might happen due to the coexistence of these languages in the same region.
    \footnote{The same could also be true for the Arabic language, as annotators identified some instances of Arabic terms in the Tamasheq speech from Section~\ref{tamasheqtrad}.}
    
    \item \textbf{Fulfulde~(FUV):} Fulfulde, also known as \textit{Fula}, \textit{Peul} or \textit{Fulani}, is a Senegambian branch of the Niger-Congo language family. Unlike most Niger-Congo languages, it does not have tones~\cite{williamson1989benue}. 
    The number of speakers is estimated to be above 40 million~\cite{hammarstrom2015ethnologue}. 
    The native speakers of this language, the Fula people, are one of the largest ethnic groups in the Sahel and West Africa~\cite{hughes2009africa}.\footnote{
    Lexical resources can be found at: \url{http://www.language-archives.org/language/ful}}
    
    \item \textbf{Hausa~(HAU):} Hausa is a Chadic language, member of the Afro-Asiatic language family. It is spoken mainly within the northern half of Nigeria and the southern half of Niger, with  \newcite{wolff1991standardization} and \newcite{newman2009hausa} estimating the number of speakers between 20 and 50 million. Early studies in Hausa showcased a remarkable number of loanwords from Arabic, Kanuri, and Tamasheq~\citelanguageresource{schon1862grammar}.


    \item \textbf{Tamasheq~(TAQ):} Tamasheq is a variety of Tuareg, a Berber macro-language spoken by nomadic tribes across North Africa in Algeria, Mali, Niger and Burkina Faso~\citelanguageresource{heath2006dictionnaire}. 
    It accounts for approximately 500,000 native speakers, being mostly spoken in Mali and Niger~\cite{hammarstrom2015ethnologue}. The livelihood of the Tuareg people has been under threat in the last century, due to climate change and a series of political conflicts~\cite{decalo1997historical}. 
    This reduced considerably the number of speakers of Tamasheq however, partially due to the Malian government's active promotion of the language in recent years, Tamasheq is now classified as a \textit{developing language}~\cite{hammarstrom2015ethnologue}.
    
    \item \textbf{Zarma~(DJE):} Zarma, also spelled \textit{Djerma}, is a leading indigenous Songhay language of the southwest lobe of the west African nation of Niger, spoken by over 2 million speakers. This tonal language is also spoken in Nigeria, Burkina Faso, Mali, Sudan, Benin and Ghana~\cite{britannica:shongailanguages}.\footnote{Lexical resources can be found at: \url{http://www.language-archives.org/language/dje}}
\end{itemize}

\section{Use case: Speech Translation Baseline}\label{baseline}

In this paper we present speech resources for the Tamasheq language, and in four other geographically close languages. They are shared in the context of the IWSLT 2022 low-resource speech translation track.
In this section we present as use case our 
end-to-end speech translation baseline that uses the Tamasheq-French Parallel Corpus from Section~\ref{tamasheqtrad}. 

\paragraph{Dataset.} We run this baseline experiment using both 
versions of the dataset from Section~\ref{tamasheqtrad}, with 
data splits detailed in Table~\ref{tab:taqfra:sets}.
We extract 80-dimensional mel filterbank features from the Tamasheq utterances. For the French text, we build a 1k unigram vocabulary using \texttt{Sentencepiece}~\cite{kudo2018sentencepiece} without pre-tokenization. 

\begin{table}
\centering
\resizebox{\columnwidth}{!}{
\begin{tabular}{cccc} \hline
\multicolumn{1}{l}{} & \textbf{train} & \textbf{valid} & \textbf{test} \\\hline
\textbf{clean (17\,h)}         & 4,444 / 13h50            & 581 / 1h53                & 804 / 1h59             \\
\textbf{full (19\,h)}         & 4,886 / 15h24            & -               & -            \\\hline
\end{tabular}}
\caption{The (Number of utterances / duration) per set. 
Both \textit{clean} and \textit{full}
share the same validation and test sets. 
}
\label{tab:taqfra:sets}
\end{table}


\paragraph{Architecture.} We use the \texttt{fairseq s2t} toolkit~\cite{wang2020fairseq}, training end-to-end speech translation Transformer models~\cite{vaswani2017attention}, preceded by two convolutional layers for dimensionality reduction.\footnote{Settings are detailed at their \texttt{s2t\_transformer\_xs} recipe.} These models are trained for 500 epochs using the Adam optimizer~\cite{kingma2014adam} with 10k warm-up steps. For decoding, we use beam search with a beam size of 5, and we evaluate the models using the best checkpoint with respect to the loss in the validation set.

\paragraph{Results and Discussion.} 
Table~\ref{tab:st_baseline} presents detokenized case-sensitive BLEU scores computed using \texttt{sacreBLEU}~\cite{post2018call}. 
Looking at these results, we notice that the \textit{full} version of the dataset improves slightly over the \textit{clean} version. The former contains roughly two extra hours in its training set, and thus 
this could hint that having more data 
in data scarcity scenarios is beneficial, even when this data is of questionable quality. 
Nevertheless, the performance of both baselines is \textit{very} low. They highlight the challenge of low-resource end-to-end speech translation when the only data used is of parallel nature. We believe results can be further improved by using auxiliary monolingual tools and models. The next paragraphs elaborate on this.

\begin{table}
\centering
\resizebox{\columnwidth}{!}{
\begin{tabular}{ccc}\hline
\textbf{}    & \textbf{valid} & \textbf{test}\\\hline
\textbf{clean~(17\,h)} & 2.22 (20.6/3.6/1.1/0.4) & 1.80 (18.8/2.9/0.8/0.3) \\
\textbf{full~(19\,h)} & 2.31 (18.5/3.3/1.0/0.4) & 1.90 (15.9/2.6/0.9/0.4) \\\hline
\end{tabular}}
\caption{End-to-end speech translation BLEU4 results for the baselines, with detailed scores between parentheses.}
\label{tab:st_baseline}
\end{table}

For the text, and since French is a high-resource language, one could incorporate pre-trained embeddings to the translation decoder. For the decoding procedure, language models -- such as \textsc{CamemBERT}~\cite{martin-etal-2020-camembert} and \textsc{FlauBERT}~\cite{le-etal-2020-flaubert-unsupervised} -- 
can be used. Pre-trained decoders like \textsc{mBART}~\cite{liu2020multilingual} could also be incorporated.

For the speech,
the self-supervised speech representation produced by models such as \textsc{HuBERT}~\cite{hsu2021hubert} and \textsc{wav2vec~2.0}~\cite{baevski2020wav2vec} 
can 
replace mel filterbanks features for the speech translation encoder. One can use freely available pretrained models in high-resource languages, or train these models from scratch. For the latter option,  the resources from Section~\ref{rawaudiocollection} can be used.
In both cases, self-supervised~(also called \textit{task-agnostic}) fine-tuning in the target language can increase results, but the best option seems to be to fine-tune in the target task directly~\cite{evain2021task,babu2021xls}.

Lastly, an interesting research direction is the leveraging of multilingual data in self-supervised models for speech. There are massive multilingual models that produce speech representations 
from many unrelated languages seen during training~\cite{conneau2020unsupervised,babu2021xls}. However, we currently do not know if these models are in fact better than having \textit{dedicated models} trained with a smaller set of languages that are closely related~(i.e. in speech style, geography, phonology, linguistic family). Thus, it might be interesting to compare the speech representations produced by a multilingual model based on the languages from Section~\ref{rawaudiocollection}, against current multilingual baselines, such as \textsc{XLSR-53}~\cite{conneau2020unsupervised} and \textsc{XLS-R}~\cite{babu2021xls}.

\section{Conclusion}\label{conclusion}

In this paper we presented two resources that focus on the Tamasheq language. The \textbf{Niger-Mali audio collection} contains 641 hours of speech in French from Niger, Fulfulde, Hausa, Tamasheq and Zarma. 
The presence in the data of audio recording from languages spoken in the same geographical area is particularly interesting for research related to transfer learning~\cite{wang2015transfer} and self-supervision~\cite{baevski2020wav2vec,hsu2021hubert,conneau2020unsupervised}.
This resource is publicly available on our website: \url{https://demo-lia.univ-avignon.fr/studios-tamani-kalangou/}.

The second resource we share with the research community, the \textbf{Tamasheq-French Parallel Corpus}, focuses on speech translation. It contains 17\,h of speech in Tamasheq 
aligned at the utterance-level to French translations.  
We believe this dataset is an interesting resource for those 
interested by low-resource speech translation.
It is publicly available 
on this \texttt{GitHub} page: ~\url{https://github.com/mzboito/IWSLT2022_Tamasheq_data}.

Lastly, we also presented a baseline model for IWSLT~2022 low-resource track using the Tamasheq-French parallel corpus. 
The obtained scores 
highlight the great challenge of developing effective approaches in such low-resource settings. We believe that by leveraging monolingual tools and data in the translation model, notably through the use of the audio collection presented in this paper, one might be able to develop more effective models for Tamasheq.

\section{Acknowledgements}
We are very thankful to Fondation Hirondelle, Studios Kalangou from Niger, and Studios Tamani from Mali, for allowing us to download, use and distribute their audio data under the Creative Commons BY-NC-ND 3.0 license for non-commercial use.\\ 
This work was funded by the French Research Agency (ANR) through the ON-TRAC project under contract number ANR-18-CE23-0021. This paper was also partially funded by the European Commission through the SELMA project under grant number 957017. We would like to thank Antoine Caubrière from LIA for all the help with the data packaging.

\section{Bibliographical References}\label{reference}

\bibliographystyle{lrec2022-bib}
\bibliography{lrec2022-example}

\section{Language Resource References}
\label{lr:ref}
\bibliographystylelanguageresource{lrec2022-bib}
\bibliographylanguageresource{languageresource}

\end{document}